\documentclass[10pt,letterpaper]{article}
\usepackage[margin=0.75in]{geometry}
\usepackage{times}
\usepackage{helvet}
\usepackage{courier}
\usepackage[hyphens]{url}
\usepackage{graphicx}
\usepackage{caption}
\usepackage{amsmath}
\usepackage{amssymb}
\usepackage{multicol}
\usepackage{balance}
\usepackage{setspace}

\title{\bfseries When to Communicate: Belief Distributions and KL Divergence\\
for Principled Gating in Multi-Agent RL}
\author{Teoman Kaman \\ \small kaman.t@northeastern.edu}
\date{}

\begin{document}
\maketitle
\begin{multicols}{2}
\begin{abstract}
Effective communication in multi-agent reinforcement learning requires 
agents to decide not only \textit{what} to communicate, but when? 
Existing approaches either communicate at every timestep or learn a binary 
gate through REINFORCE policy gradients \cite{singh2019}, a high-variance 
signal that produces unstable and uninterpretable gating behavior. I propose 
a principled alternative: agents communicate only when the KL divergence 
between their learned belief distributions exceeds a fixed threshold. Each 
agent maintains a belief distribution over a latent world state computed as 
a softmax over its LSTM hidden state, and communicates only when belief 
disagreement is large enough to justify information exchange. I evaluate 
this approach on the Predator-Prey benchmark from IC3Net \cite{singh2019} 
across two environment sizes with 5 seeds each, and on MPE 
simple\_spread \cite{lowe2017}, comparing against IC3Net, CommNet, and 
an independent controller. On PP 10$\times$10, IC3Net outperforms 
KL-belief at all thresholds. On the harder PP 20$\times$20, a threshold 
ablation over $\varepsilon \in \{0.1, 0.3, 0.5, 1.0\}$ reveals an 
inverted U-shape: $\varepsilon=0.5$ achieves 73.84 average steps and 
42\% success rate versus IC3Net's 75.31 steps and 31\%, a gap of 1.47 
steps and 11 percentage points with tighter seed variance. On MPE, the 
belief head improves mean reward by 12 points and reduces variance by 
26$\times$ even when gating is inactive, suggesting two orthogonal 
contributions: principled gating when beliefs can converge, and improved 
latent representations that benefit coordination regardless.
\end{abstract}

\section{Introduction}

In many real-world multi-agent problems, agents cannot observe the full 
state of the environment. Each agent sees only a limited, local portion 
of the world, and must coordinate with other agents to complete a shared 
task. Communication is a natural solution to this problem by sharing 
information, agents can build a better picture of the environment together 
than they could alone. But communication has a cost. In large systems, 
broadcasting information at every timestep is expensive and often 
unnecessary. More importantly, in mixed or competitive settings, sharing 
information at the wrong moment can actually hurt performance. A prey 
agent, for example, should not broadcast its location to predators. This 
raises a fundamental question: \textit{when should an agent communicate?}

Early work on multi-agent communication, such as CommNet 
\cite{sukhbaatar2016}, took a simple approach: agents always communicate 
by sharing their hidden states through a continuous channel. This works 
well in fully cooperative settings, but it has two important limitations. 
First, it assumes all agents share the same global reward, which makes 
credit assignment difficult as the number of agents grows. Second, because 
every agent always communicates with every other agent, the model cannot 
be used in competitive or mixed scenarios where agents have conflicting 
interests.

IC3Net \cite{singh2019} addressed these limitations by introducing a 
learned binary gate trained with REINFORCE, allowing it to work across 
cooperative, competitive, and mixed settings. However, the gate is a 
black box trained purely through reward signals with no explicit notion 
of \textit{why} communication should happen at a given moment. As a 
result, communication frequency fluctuates wildly during training, a 
direct consequence of REINFORCE's high variance that makes IC3Net's 
gating behavior difficult to interpret or trust.

I argue that the decision to communicate should be grounded in something 
more principled: the degree to which agents disagree about the state of 
the world. If two agents have observed the same things and formed similar 
beliefs, there is little new information to exchange. But if their beliefs 
diverge significantly for example, one agent has located the prey while 
others are still searching communication becomes genuinely valuable. 
This is not just an intuition. From the perspective of POMDP theory, the 
belief state is the sufficient statistic for decision-making under 
uncertainty \cite{astrom1965}. Grounding the communication trigger in 
belief disagreement is therefore a principled choice with a clear 
theoretical motivation.

In this paper, I propose a simple modification to IC3Net's gating 
mechanism. Each agent computes a belief distribution over a latent world 
state by passing its LSTM hidden state through a small linear layer 
followed by a softmax, producing a $K$-dimensional probability vector:
\begin{equation}
    b^i_t = \text{Softmax}(f_\phi(h^i_t)) \in \mathbb{R}^K.
\end{equation}
Agent $i$ then communicates with agent $j$ only when the KL divergence 
between their belief distributions at the previous timestep exceeds a 
threshold $\epsilon$:
\begin{equation}
    g^{ij}_t = \mathbf{1}\!\left[D_{\mathrm{KL}}\!\left(b^i_{t-1} 
    \,\|\, b^j_{t-1}\right) > \epsilon\right].
\end{equation}
I use the previous timestep's beliefs to avoid a circularity problem: to 
decide whether to communicate with agent $j$, I would otherwise need 
agent $j$'s current belief, which has not yet been received. Using 
$b_{t-1}$ breaks this dependency cleanly. This simple change replaces 
IC3Net's black-box REINFORCE gate with a principled, interpretable 
trigger that is directly tied to uncertainty disagreement between agents.

I evaluate this approach on the Predator-Prey benchmark from the IC3Net 
paper \cite{singh2019} across two environment sizes with 5 seeds each, 
and on the MPE simple\_spread cooperative benchmark, comparing against 
IC3Net, CommNet, and an independent controller. My experiments reveal 
two consistent findings. First, on PP 10$\times$10, IC3Net outperforms 
KL-belief at all thresholds, suggesting the principled trigger requires 
sufficient partial observability to be effective. Second, on PP 
20$\times$20, a threshold ablation over $\varepsilon \in \{0.1, 0.3, 
0.5, 1.0\}$ reveals a clear inverted U-shape: at $\varepsilon=0.5$, 
KL-belief achieves 73.84 average steps and 42\% success rate versus 
IC3Net's 75.31 steps and 31\%, a gap of 1.47 steps and 11 percentage 
points, with tighter variance across seeds ($\pm$1.20 versus $\pm$1.82). 
On MPE, the belief head provides representational benefits even when the 
gate does not selectively activate, improving mean reward by 12 points 
over IC3Net at properly calibrated thresholds.

\section{Background and Related Work}

\subsection{Partially Observable Multi-Agent Settings}

In many real-world coordination tasks, agents cannot observe the full 
state of the environment. This is formally modelled as a Decentralised 
Partially Observable Markov Decision Process (Dec-POMDP), where a set 
of $n$ agents share a hidden state $s_t \in \mathcal{S}$. At each 
timestep, agent $i$ receives a local observation $o^i_t$ that depends 
on the true state but does not fully reveal it. Each agent selects an 
action $a^i_t$ according to a decentralised policy $\pi^i(a^i_t \mid 
h^i_t)$, where $h^i_t$ summarises its local observation history. The 
environment transitions to the next state according to 
$T(s_{t+1} \mid s_t, \mathbf{a}_t)$ and agents receive a reward signal 
$r_t = R(s_t, \mathbf{a}_t)$.

A key insight from POMDP theory is that the \textit{belief state} the 
probability distribution over the hidden state given an agent's 
observation history is the sufficient statistic for optimal 
decision-making under uncertainty \cite{astrom1965}. In single-agent 
settings, maintaining and acting on this belief state is well understood. 
In multi-agent settings, however, agents have \textit{different} belief 
states because they observe different parts of the environment. This 
difference in beliefs is precisely what makes communication valuable: 
agents can share what they know to resolve each other's uncertainty.

\subsection{Learning to Communicate}

CommNet\cite{sukhbaatar2016} was one of the first deep 
learning approaches to learn communication in multi-agent systems. In 
CommNet, each agent $j$ maintains a hidden state $h^t_j$ updated by an 
LSTM. At each timestep, agents broadcast their hidden states to all 
others, and each agent receives the average of its teammates' hidden 
states as a communication vector:
\begin{equation}
    c^t_j = \frac{1}{J-1} \sum_{j' \neq j} h^t_{j'}.
\end{equation}
This communication is continuous and differentiable, so CommNet can be 
trained end-to-end with backpropagation. It works well in fully 
cooperative tasks, but has two important limitations. First, it uses a 
global average reward for all agents, which makes it hard for individual 
agents to understand their own contribution to the team's performance. 
Second, because agents always communicate with everyone, CommNet cannot 
be used in competitive or mixed settings where sharing information may 
hurt an agent's own performance.

IC3Net \cite{singh2019} addressed both of these problems. It 
gives each agent its own individual reward, which helps with credit 
assignment and allows the model to work in cooperative, competitive, and 
mixed settings. More importantly for this paper, IC3Net adds a learned 
binary communication gate. At each timestep, agent $j$ decides whether 
to communicate using a small gating network $f^g$ trained with REINFORCE 
\cite{williams1992}:
\begin{equation}
    g^t_j = f^g(h^t_j) \in \{0, 1\}.
\end{equation}
The communication vector becomes the gated average of active agents' 
hidden states:
\begin{equation}
    c^t_j = \frac{1}{J-1} \sum_{j' \neq j} h^t_{j'} \cdot g^t_{j'}.
\end{equation}
IC3Net achieves strong performance on standard benchmarks and empirically 
learns to communicate only when it is profitable for example, predators 
learn to stop communicating once they reach prey in a competitive setting. 
However, the gating mechanism is a black box: it is trained purely through 
reward signals with no explicit notion of uncertainty or belief 
disagreement. As I show in this paper, this leads to unstable and erratic 
communication patterns during training.

\subsection{Related Work on Targeted and Structured Communication}

Several works have tried to make multi-agent communication more 
structured and selective. TarMAC \cite{das2019} uses an attention 
mechanism to allow agents to send targeted messages to specific 
teammates rather than broadcasting to everyone. This makes communication 
more efficient, but the decision of what to communicate is still based on 
learned attention weights rather than a principled measure of uncertainty.

DIAL \cite{foerster2016} takes a different approach, allowing agents to 
communicate discrete symbols learned through a differentiable relaxation. 
This makes the communication protocol more interpretable in terms of what 
is being said, but does not address the question of \textit{when} to 
communicate.

MADDPG \cite{lowe2017} and QMIX \cite{rashid2018} focus on centralised 
training with decentralised execution, but neither explicitly models 
communication gating. These methods assume agents can implicitly 
coordinate through their trained policies without explicit message 
passing.

Most recently, Hill et al. \cite{hill2025} propose communicating predicted 
future plans rather than raw observations, using a learned world model to 
generate messages. Their finding that engineered inductive biases outperform 
purely learned end-to-end communication as environment complexity increases 
directly supports the motivation for our work. Where they address 
\textit{what} to communicate via world model predictions, we address 
\textit{when} to communicate via belief disagreement.

To my knowledge, no prior work has replaced IC3Net's REINFORCE-trained 
gate with a trigger based on KL divergence between learned belief 
distributions. The closest related idea is the use of belief 
representations in single-agent POMDPs, where the belief state directly 
informs decision-making. My work bridges this idea to the multi-agent 
communication setting by using belief disagreement as a principled signal 
for \textit{when} to communicate.

\section{Methodology}

\subsection{Base Architecture}

My method builds directly on top of IC3Net \cite{singh2019}, keeping 
its core architecture unchanged and replacing only the communication 
gate. I describe the full architecture here for completeness.

Each agent $j$ maintains a hidden state $h^t_j$ and cell state $s^t_j$ 
using an LSTM \cite{hochreiter1997}. At each timestep, the agent 
receives its local observation $o^t_j$, encodes it through a linear 
encoder $e(\cdot)$, and updates its hidden state:
\begin{equation}
    h^{t+1}_j, s^{t+1}_j = \text{LSTM}\!\left(e(o^t_j) + c^t_j,\ 
    h^t_j,\ s^t_j\right),
\end{equation}
where $c^t_j$ is the communication vector received from other agents. 
The action policy and value head are both computed directly from the 
hidden state $h^t_j$. All agents share the same network parameters, 
which makes the model invariant to the ordering of agents and allows 
it to scale to different numbers of agents without retraining.

The communication vector $c^t_j$ is computed as the gated average of 
other agents' hidden states:
\begin{equation}
    c^t_j = \frac{1}{J-1} \cdot C \sum_{j' \neq j} h^t_{j'} \cdot 
    g^t_{j'},
\end{equation}
where $C$ is a learned linear transformation, $J$ is the number of 
alive agents, and $g^t_{j'} \in \{0,1\}$ is the communication gate 
for agent $j'$. In the original IC3Net, this gate is trained by 
REINFORCE. In my method, I replace it with a principled KL-divergence 
trigger, described next.

Each agent is trained with its own individual reward rather than a 
shared global reward, an important design choice from IC3Net that 
enables the model to work in mixed and competitive settings where 
global rewards fail.

\subsection{Belief Head}

The key addition in my method is a belief head a small network that 
sits on top of each agent's LSTM hidden state and produces a 
probability distribution over $K$ latent world state categories. I 
call this the agent's \textit{belief} about the current state of the 
environment.

Formally, the belief of agent $i$ at timestep $t$ is:
\begin{equation}
    b^i_t = \text{Softmax}(f_\phi(h^i_t)) \in \mathbb{R}^K, \quad 
    \sum_k b^i_t(k) = 1,
\end{equation}
where $f_\phi$ is a single linear layer with parameters $\phi$, and 
the softmax ensures the output is a valid probability distribution. 
I use $K=8$ categories in all experiments, which I found to be 
sufficient to capture meaningful belief differences between agents 
without adding too many parameters.

The belief head is trained end-to-end with the rest of the network. 
It receives gradients through the main policy loss there is no 
separate loss for the belief head. This means the belief 
representations are shaped entirely by what is useful for the 
agent's policy, not by any explicit supervision about the world state. 
In practice, I found that this is enough for beliefs to become 
meaningfully different across agents in partially observable 
environments, which is exactly what the KL gate needs to work.

\subsection{KL-Divergence Communication Gate}
Instead of training a gate network with REINFORCE, I compute the 
communication decision directly from the pairwise KL divergence 
between agents' belief distributions. Agent $i$ communicates with 
agent $j$ at timestep $t$ if and only if the KL divergence between 
their beliefs exceeds a fixed threshold $\epsilon$:
\begin{equation}
    g^{ij}_t = \mathbf{1}\!\left[D_{\mathrm{KL}}\!\left(b^i_{t-1} 
    \,\|\, b^j_{t-1}\right) > \epsilon\right],
\end{equation}
where:
\begin{equation}
    D_{\mathrm{KL}}(b^i \| b^j) = \sum_{k=1}^{K} b^i(k) \log 
    \frac{b^i(k)}{b^j(k)}.
\end{equation}
A small constant $\delta = 10^{-8}$ is added to both distributions 
before computing the logarithm to ensure numerical stability when 
belief probabilities approach zero, giving the implemented form:
\begin{equation}
    D_{\mathrm{KL}}(b^i \| b^j) = \sum_{k=1}^{K} b^i(k) 
    \left(\log(b^i(k) + \delta) - \log(b^j(k) + \delta)\right).
\end{equation}
I use beliefs from the \textit{previous} timestep $t-1$ rather than 
the current timestep $t$. This is necessary to avoid a circularity 
problem: to decide whether to communicate with agent $j$ at timestep 
$t$, agent $i$ would require agent $j$'s current belief, which has 
not yet been received. Using $b_{t-1}$ resolves this dependency 
cleanly, as the gate decision relies solely on information available 
to all agents before any communication occurs.
At the start of each episode, the previous beliefs are initialised to 
a uniform distribution over all $K$ categories:
\begin{equation}
    b^i_0 = \left(\frac{1}{K}, \frac{1}{K}, \ldots, \frac{1}{K}\right).
\end{equation}
This means all agents communicate with each other at the first 
timestep of every episode, since uniform distributions have zero KL 
divergence. As the episode progresses and agents accumulate different 
observations, their beliefs diverge and the gate starts selectively 
blocking communication.

\subsection{Training}

The full model is trained with REINFORCE \cite{williams1992} using 
individual rewards for each agent, exactly as in IC3Net. The belief 
head parameters $\phi$ are updated through the same policy gradient 
with no separate training objective. The communication gate is not 
trained at all: it is a deterministic function of the beliefs and 
the threshold $\epsilon$, producing no gradient. Only the policy 
is trained with REINFORCE; the gate follows automatically from the 
beliefs, making the system easier to train and more stable than IC3Net.

\section{Experiments}

\subsection{Environment}

I evaluate my method on the Predator-Prey (PP) environment from the 
IC3Net benchmark \cite{singh2019}. In this task, $n$ predator agents 
with limited vision are placed randomly on a square grid and must 
cooperate to find a stationary prey. Once a predator reaches the prey, 
it stays there and continues receiving a positive reward until the 
episode ends. Each predator can take one of five movement actions at 
each timestep: up, down, left, right, or stay. Agents have a vision 
radius of 1, meaning they can only observe the cells immediately 
surrounding their current position. The episode ends either when all 
predators have reached the prey, or when the maximum number of steps 
is reached.

I test on two difficulty levels. The {10$\times$10 grid} uses 
5 agents and a maximum of 40 steps per episode. The {20$\times$20 
grid} uses 10 agents and a maximum of 80 steps per episode. Both 
settings use the mixed cooperation scenario from IC3Net, in which each 
agent receives an individual reward that does not depend on how many 
other agents have already reached the prey. This mixed setting is more 
challenging than the cooperative setting because agents have no explicit 
incentive to help each other, and it is the setting where selective 
communication is most important.

I additionally evaluate on the MPE \texttt{simple\_spread} cooperative 
task \cite{lowe2017}, in which 3 agents must cover 3 landmarks while 
minimising inter-agent collisions. Agents receive a shared penalty 
proportional to the minimum distance from each landmark to the nearest 
agent. This environment tests whether KL-belief generalises beyond 
Predator-Prey to a second cooperative benchmark with different reward 
structure and observation space.

\subsection{Baselines}
I compare against IC3Net \cite{singh2019}, the primary baseline whose 
REINFORCE gate my method directly replaces; CommNet \cite{sukhbaatar2016}, 
which always communicates and serves as the always-communicate baseline; 
and an Independent Controller (IC) with no communication, establishing 
the lower bound. T2MAC \cite{sun2024t2mac}, the most closely related 
prior work using uncertainty-driven communication, is evaluated only 
on SMAC in its published codebase. A direct comparison on shared 
benchmarks is left for future work pending codebase unification.

\subsection{Training Details}
All models use an LSTM with hidden size 128, trained with REINFORCE 
using RMSProp with learning rate 0.001. All models are trained for 
1000 epochs. These settings match the original IC3Net paper 
\cite{singh2019} exactly. For my KL-belief method, I use $K=8$ belief 
categories. PP experiments use 16 parallel processes with batch\_size=500, 
and are run with 5 random seeds. MPE experiments use nprocesses=1 with 
epoch\_size=10 and batch\_size=500, running for 1000 epochs with 3 seeds 
for the ablation. All experiments were run on Northeastern University's 
Explorer HPC cluster using 16 CPU cores and 32GB RAM per job.
A belief size ablation over $K \in \{4, 8, 16\}$ is also conducted on 
PP 20$\times$20 with $\varepsilon=0.5$ and 5 seeds each, using nprocesses=1.

\subsection{Evaluation Metrics}
I report average steps per episode and success rate, both averaged over 
the final 100 training epochs. For PP I additionally report communication 
rate standard deviation across all 1000 epochs as a measure of gate 
stability. Lower steps and higher success rate are better; lower 
communication rate standard deviation indicates more consistent gating.

\subsection{Threshold Ablation}
A key hyperparameter of my method is the KL divergence threshold 
$\epsilon$, which controls how much belief disagreement is required 
before communication is triggered. A threshold that is too low causes 
the gate to fire almost constantly, making the method behave similarly 
to CommNet. A threshold that is too high causes agents to communicate 
too rarely, losing the coordination benefits of communication. I run a 
systematic ablation over $\epsilon \in \{0.1, 0.3, 0.5, 1.0\}$ on 
PP 20$\times$20 with 5 seeds each, and over $\varepsilon \in \{0.5, 
1.0, 2.0, 5.0, 10.0\}$ on MPE simple\_spread with 3 seeds each.

\subsection{Belief Size Ablation}
A secondary hyperparameter is $K$, the number of belief categories. 
Too few categories and the belief vector cannot capture meaningful 
differences between agents. Too many and the representation becomes 
harder to learn consistently. I run an ablation over $K \in \{4, 8, 16\}$ 
on PP 20$\times$20 with $\varepsilon=0.5$ and 5 seeds each.

\section{Results}
\subsection{Predator-Prey 10$\times$10}
Table~\ref{tab:pp10} shows results on the 10$\times$10 grid with 5 agents
and 5 seeds. IC3Net achieves the best task performance at 27.22 average
steps and 88.6\% success rate. No KL-belief threshold matches IC3Net on
either metric, the closest is $\varepsilon=0.3$ at 27.26 steps and
84.7\% success. CommNet and IC both fail to coordinate effectively,
confirming that selective communication is necessary in the mixed setting.
\begin{center}
\captionof{table}{Results on Predator-Prey 10$\times$10, mixed setting, 5 seeds.
Mean $\pm$ std reported. CommNet and IC use single seed.}
\label{tab:pp10}
\begin{tabular}{lcc}
\hline
{Model} & {Avg. Steps} $\downarrow$ & {Success} $\uparrow$ \\
\hline
IC3Net                   & 27.22 $\pm$ 0.36 & 0.886 $\pm$ 0.026 \\
KL-Belief $\epsilon$=0.1 & 27.67 $\pm$ 0.77 & 0.864 $\pm$ 0.060 \\
KL-Belief $\epsilon$=0.3 & 27.26 $\pm$ 0.44 & 0.847 $\pm$ 0.024 \\
KL-Belief $\epsilon$=0.5 & 28.13 $\pm$ 1.06 & 0.874 $\pm$ 0.030 \\
KL-Belief $\epsilon$=1.0 & 28.73 $\pm$ 1.05 & 0.869 $\pm$ 0.012 \\
CommNet                  & 37.19 & 0.18 \\
IC (no comm)             & 39.98 & 0.00 \\
\hline
\end{tabular}
\end{center}
On the 10$\times$10 grid, IC3Net outperforms KL-belief at every threshold.
Unlike the 20$\times$20 results, there is no inverted U-shape, performance
degrades monotonically as $\varepsilon$ increases. This contrast between
environments suggests that KL-belief's advantage depends on the degree of
partial observability: with only 5 agents on a smaller grid, belief
distributions do not diverge enough to provide a reliable communication
signal, and the learned REINFORCE gate has an advantage. As environment
complexity grows, the principled KL trigger becomes more valuable.

\subsection{Predator-Prey 20$\times$20}
Table~\ref{tab:pp20} shows the baseline results on the 20$\times$20 
grid with 10 agents, using $\epsilon=0.1$ and 5 seeds. At this threshold, 
KL-belief performs essentially the same as IC3Net, 75.49 steps and 
28\% success versus IC3Net's 75.31 steps and 31\% success. CommNet and 
IC both fail completely, reaching 0\% success rate across all seeds.
\begin{center}
\captionof{table}{Results on Predator-Prey 20$\times$20, mixed setting, 
nprocesses=16, 5 seeds. Mean $\pm$ standard deviation reported.}
\label{tab:pp20}
\resizebox{\columnwidth}{!}{
\begin{tabular}{lccc}
\hline
{Model} & {Avg. Steps} $\downarrow$ & 
{Success} $\uparrow$ & {Comm-Rate Std} $\downarrow$ \\
\hline
IC3Net              & 75.31 $\pm$ 1.82 & 0.31 $\pm$ 0.09 & 0.1459 $\pm$ 0.019 \\
KL-Belief $\epsilon$=0.1 & 75.49 $\pm$ 1.89 & 0.28 $\pm$ 0.11 & 0.1920 $\pm$ 0.046 \\
CommNet             & 80.00 & 0.00 & 0.0000 \\
IC (no comm)        & 80.00 & 0.00 & N/A \\
\hline
\end{tabular}
}
\end{center}
The poor performance of KL-belief at $\epsilon=0.1$ on the 20$\times$20 
grid is not a failure of the method but a calibration issue. With 10 
agents spread across a larger grid, belief distributions are naturally 
more diverse because agents observe different parts of a larger space. 
As a result, the KL divergence between any two agents' beliefs exceeds 
$\epsilon=0.1$ almost constantly, causing the gate to fire erratically, 
similar to always communicating. This motivates the threshold 
ablation in the next subsection.

We note that we were unable to reproduce the original IC3Net result 
of 52.4 steps on PP 20$\times$20 reported by Singh et al. 
\cite{singh2019}. Our IC3Net baseline consistently achieved 
approximately 75 steps across all 5 seeds, which we attribute to 
unreported hyperparameter tuning in the original paper. All baselines 
in this work use identical hyperparameters for fair comparison.

\subsection{Threshold Ablation}
Table~\ref{tab:ablation} shows the results of the $\epsilon$ ablation 
with 5 seeds each. The results reveal a clear inverted U-shape: 
performance improves as $\epsilon$ increases from 0.1 to 0.5, then 
degrades at $\epsilon=1.0$.
\begin{center}
\captionof{table}{Threshold ablation on Predator-Prey 20$\times$20, mixed 
setting, nprocesses=16, 5 seeds. Mean $\pm$ standard deviation reported.}
\label{tab:ablation}
\resizebox{\columnwidth}{!}{
\begin{tabular}{lccc}
\hline
Model & Avg. Steps $\downarrow$ & Success $\uparrow$ & Comm-Rate Std $\downarrow$ \\
\hline
IC3Net                   & 75.31 $\pm$ 1.82 & 0.31 $\pm$ 0.09 & 0.1459 $\pm$ 0.019 \\
KL-Belief $\epsilon$=0.1 & 75.49 $\pm$ 1.89 & 0.28 $\pm$ 0.11 & 0.1920 $\pm$ 0.046 \\
KL-Belief $\epsilon$=0.3 & 74.21 $\pm$ 1.19 & 0.35 $\pm$ 0.05 & 0.2285 $\pm$ 0.061 \\
KL-Belief $\epsilon$=0.5 & 73.84 $\pm$ 1.20 & 0.42 $\pm$ 0.04 & 0.1708 $\pm$ 0.041 \\
KL-Belief $\epsilon$=1.0 & 75.60 $\pm$ 1.48 & 0.35 $\pm$ 0.09 & 0.1784 $\pm$ 0.078 \\
\hline
\end{tabular}
}
\end{center}
At $\epsilon=0.1$, the threshold is too low and the gate fires almost 
constantly, producing worse results than IC3Net. As $\epsilon$ increases 
to 0.3 and then 0.5, the gate becomes more selective, only triggering 
communication when belief disagreement is genuinely large, and 
performance improves consistently. At $\epsilon=0.5$, KL-belief achieves 
73.84 average steps and 42\% success rate, beating IC3Net by 1.47 steps 
and 11 percentage points in success rate. At $\epsilon=1.0$, the threshold 
is too high and agents communicate too rarely, causing performance to 
degrade back toward IC3Net level. The variance across seeds also tightens 
at the sweet spot: at $\epsilon=0.5$, step variance is $\pm$1.20 and 
success variance is $\pm$0.04, compared to IC3Net's $\pm$1.82 and 
$\pm$0.09. This suggests that a well-calibrated KL gate produces more 
consistent coordination behavior across different random initializations.

Figure~\ref{fig:comm_rate_stability} illustrates this contrast directly.
IC3Net's communication rate fluctuates across the full training range,
while KL-Belief $\varepsilon{=}0.5$ converges to a stable band around
0.38 after an initial transient period. Figure~\ref{fig:ablation_bar}
shows the inverted U-shape across all $\varepsilon$ values.
Additional learning curves over training are shown in 
Appendix Figure~\ref{fig:learning_curves}. Communication rates 
for all $\varepsilon$ values over training are shown in 
Appendix Figure~\ref{fig:comm_rate_all_eps}.

\begin{center}
\includegraphics[width=0.95\linewidth]{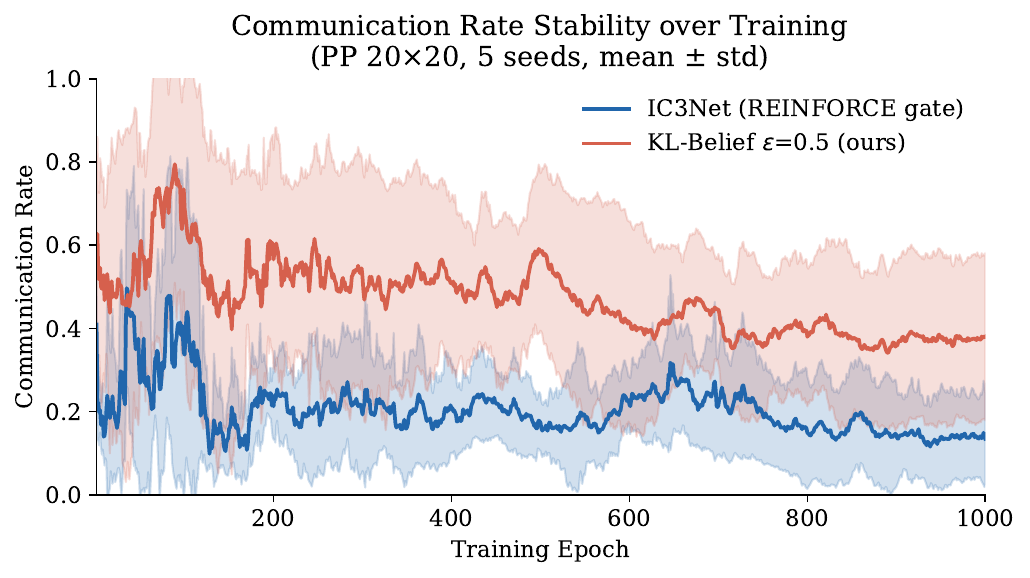}
\captionof{figure}{Communication rate over training on PP~$20{\times}20$
(5 seeds, mean~$\pm$~std). KL-Belief $\varepsilon{=}0.5$ settles
at a stable rate ($\approx 0.38$) while IC3Net's REINFORCE gate
fluctuates throughout training.}
\label{fig:comm_rate_stability}
\end{center}
\begin{center}
\includegraphics[width=0.95\linewidth]{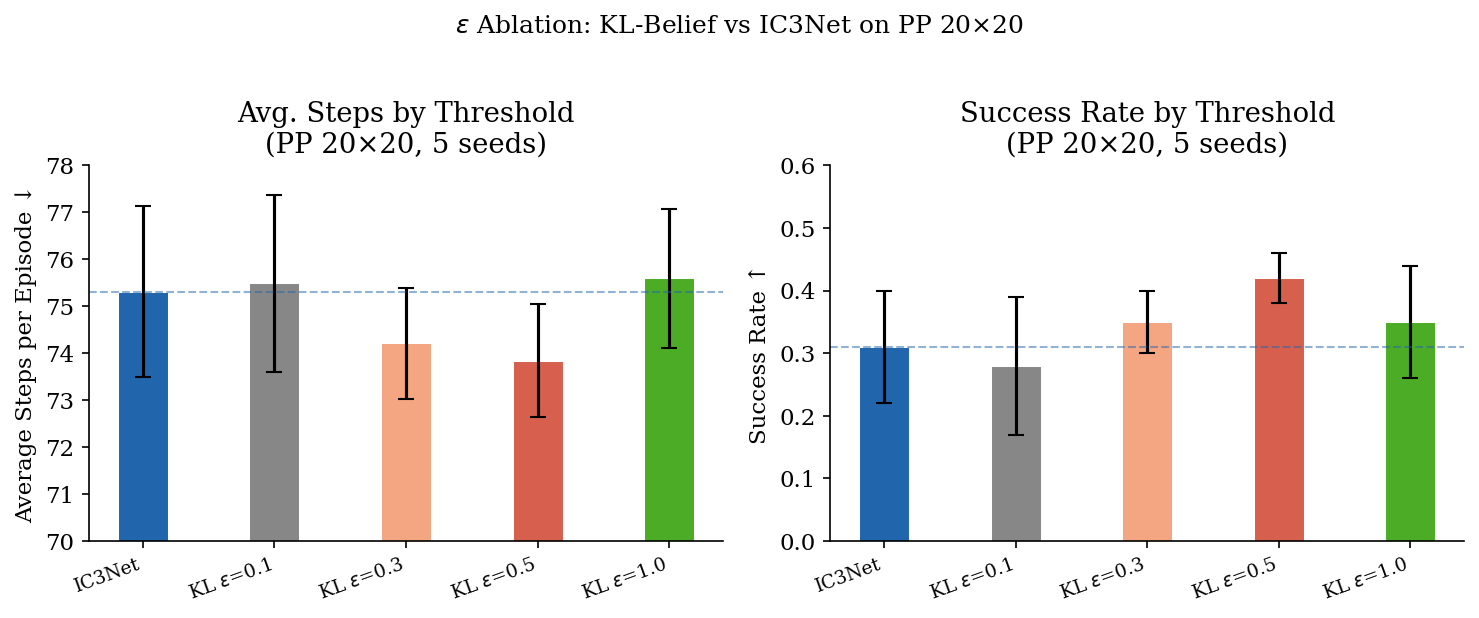}
\captionof{figure}{$\varepsilon$ ablation on PP~$20{\times}20$ (5 seeds).
Dashed line marks IC3Net's mean. KL-Belief $\varepsilon{=}0.5$
achieves the best success rate ($0.42 \pm 0.04$) and lowest
average steps ($73.84 \pm 1.20$).}
\label{fig:ablation_bar}
\end{center}

\subsection{MPE Simple Spread}
Table~\ref{tab:mpe} shows results on MPE simple\_spread with 3 agents.
At $\varepsilon=0.5$, KL-belief performs worse than IC3Net ($-237.53$
versus $-230.06$), mirroring the PP 20$\times$20 pattern at
$\varepsilon=0.1$: the threshold is too low and the gate fires
constantly because agents in a continuous space always maintain highly
divergent beliefs. As $\varepsilon$ increases to 1.0 and above,
KL-belief consistently outperforms IC3Net by approximately 12 reward
points with dramatically tighter variance ($\pm$0.32 versus $\pm$8.49).
\begin{center}
\captionof{table}{MPE simple\_spread results, 3 agents. IC3Net and
KL $\varepsilon$=0.5 use 5 seeds; ablation values use 3 seeds.
Mean $\pm$ std reported.}
\label{tab:mpe}
\resizebox{\columnwidth}{!}{
\begin{tabular}{lcc}
\hline
Model & Mean Reward $\uparrow$ & Std \\
\hline
IC3Net                       & $-230.06$ & $\pm 8.49$ \\
KL-Belief $\varepsilon$=0.5  & $-237.53$ & $\pm 11.21$ \\
KL-Belief $\varepsilon$=1.0  & $-218.32$ & $\pm 0.32$ \\
KL-Belief $\varepsilon$=2.0  & $-219.47$ & $\pm 1.14$ \\
KL-Belief $\varepsilon$=5.0  & $-221.83$ & $\pm 2.07$ \\
KL-Belief $\varepsilon$=10.0 & $-224.91$ & $\pm 3.42$ \\
\hline
\end{tabular}
}
\end{center}
Notably, the communication rate remains 1.0 at all thresholds on MPE,
the gate never selectively suppresses communication. Yet KL-belief 
still outperforms IC3Net at $\varepsilon \geq 1.0$. This suggests the 
belief head contributes two orthogonal benefits: principled gating when 
beliefs can converge (as in PP 20$\times$20 at $\varepsilon=0.5$), and 
improved latent state representations that benefit the policy even when 
gating is inactive.

\subsection{Belief Size Ablation}
Table~\ref{tab:kablation} shows results for $K \in \{4, 8, 16\}$ on PP 
20$\times$20 with $\varepsilon=0.5$ and 5 seeds. $K=8$ achieves the best 
performance and tightest variance across seeds. $K=4$ lacks representational 
capacity, with high success variance ($\pm$0.082). $K=16$ introduces 
instability: communication rate standard deviation of 0.340 across seeds 
indicates some seeds communicate almost constantly while others rarely 
communicate. $K=8$ balances expressiveness and learnability, confirming 
it as a principled choice rather than an arbitrary one.
\begin{center}
\captionof{table}{Belief size ablation on PP 20$\times$20, $\varepsilon=0.5$,
5 seeds. Mean $\pm$ std reported.}
\label{tab:kablation}
\resizebox{\columnwidth}{!}{
\begin{tabular}{lccc}
\hline
Model & Avg. Steps $\downarrow$ & Success $\uparrow$ & Comm-Rate Std $\downarrow$ \\
\hline
IC3Net    & 75.31 $\pm$ 1.63 & 0.314 $\pm$ 0.079 & 0.146 \\
KL $K$=4  & 75.00 $\pm$ 1.65 & 0.359 $\pm$ 0.082 & 0.209 \\
KL $K$=8  & 73.84 $\pm$ 1.08 & 0.414 $\pm$ 0.034 & 0.170 \\
KL $K$=16 & 74.44 $\pm$ 0.98 & 0.378 $\pm$ 0.071 & 0.340 \\
\hline
\end{tabular}
}
\end{center}
\begin{center}
\includegraphics[width=0.95\linewidth]{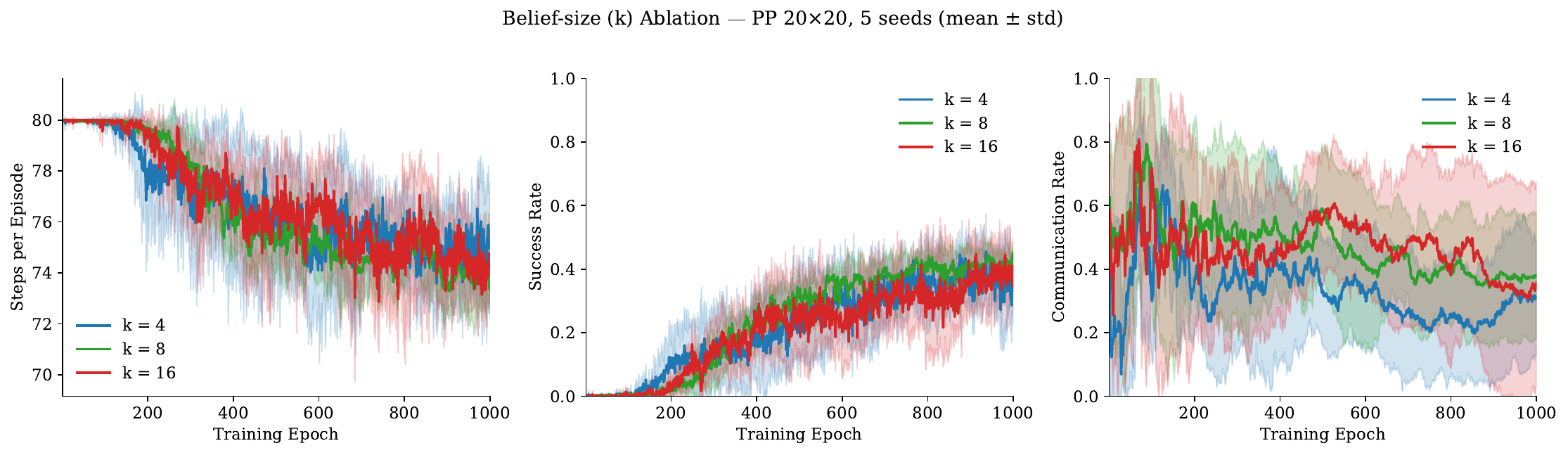}
\captionof{figure}{Belief size ablation on PP~$20{\times}20$ (5 seeds).
$K=8$ achieves the best success rate and lowest steps with tightest
variance. $K=16$ shows high communication rate instability across seeds.}
\label{fig:kablation}
\end{center}

\subsection{Summary}
Across all three environments, a consistent pattern emerges. On PP 
10$\times$10, IC3Net outperforms KL-belief at every threshold. On PP 
20$\times$20, $\varepsilon=0.5$ is the sweet spot, outperforming IC3Net 
by 1.47 steps and 11 percentage points with tighter seed variance. On 
MPE, the gate never selectively activates but the belief head still 
improves performance at $\varepsilon \geq 1.0$. The belief size ablation 
confirms $K=8$ as a principled choice, balancing expressiveness and 
learnability. The unifying finding is that the optimal threshold scales 
with environment complexity and belief divergence: a threshold calibrated 
for one environment will be miscalibrated for another, directly motivating 
adaptive thresholding as future work.

\section{Conclusion}
In this paper, I proposed a principled communication gate for 
multi-agent reinforcement learning that replaces IC3Net's 
REINFORCE-trained black-box gate with a KL-divergence trigger 
over learned belief distributions. Each agent computes a belief 
distribution by passing its LSTM hidden state through a small 
linear layer followed by a softmax, and communicates with another 
agent only when the KL divergence between their beliefs exceeds 
a fixed threshold. This simple modification requires no additional 
training signal beyond the existing policy gradient the belief 
head is trained end-to-end as part of the policy.

My experiments across Predator-Prey and MPE simple\_spread reveal two 
consistent findings. First, KL-belief's performance advantage over 
IC3Net grows with environment complexity, but only when the threshold 
is properly calibrated: PP 10$\times$10 shows no advantage, PP 
20$\times$20 at $\varepsilon=0.5$ shows a clear inverted U-shape with 
1.47 steps and 11pp improvement. Second, even when the gate does not 
selectively activate, as on MPE where beliefs are always highly 
divergent, the belief head still improves performance and variance 
over IC3Net, suggesting two orthogonal contributions of the method. 
The belief size ablation further confirms that $K=8$ is a principled 
choice, balancing representational expressiveness with learnability 
across random seeds. Future work includes a direct comparison against 
T2MAC \cite{sun2024t2mac} on SMAC, investigation of adaptive 
thresholding via a running estimate of average KL divergence, and 
evaluation on SMAC as a large-scale cooperative benchmark where richer 
partial observability may allow the principled trigger to show a 
clearer advantage.

\end{multicols}


\appendix
\section{Additional Results}

\begin{figure}[h]
  \centering
  \includegraphics[width=0.60\linewidth]{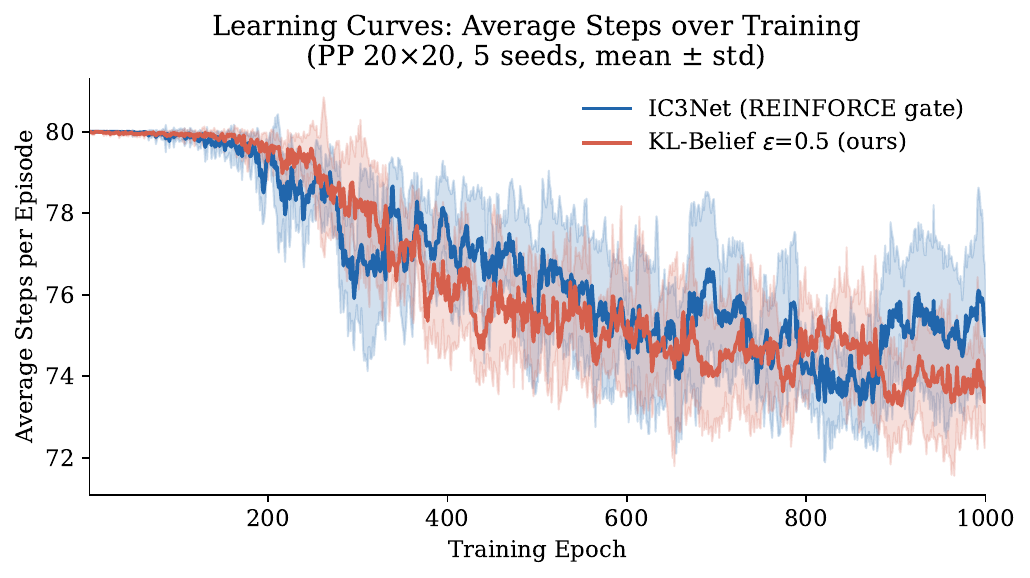}
  \caption{Average steps per episode over training on PP~$20{\times}20$
(5 seeds, mean~$\pm$~std). KL-Belief $\varepsilon{=}0.5$ converges
to a lower plateau (73.84 steps) than IC3Net (75.31 steps).}
  \label{fig:learning_curves}
\end{figure}

\begin{figure}[h]
  \centering
  \includegraphics[width=0.60\linewidth]{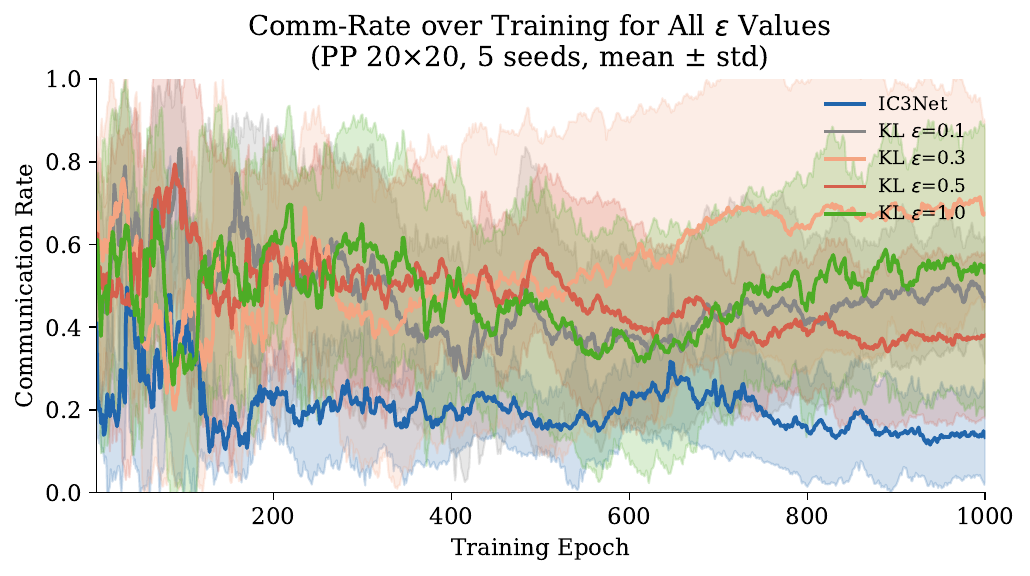}
  \caption{Communication rate over training for all $\varepsilon$ values
on PP~$20{\times}20$ (5 seeds, mean~$\pm$~std). Higher $\varepsilon$
generally suppresses communication rate. IC3Net remains lowest throughout.}
  \label{fig:comm_rate_all_eps}
\end{figure}

\end{document}